# Disaster Monitoring using Unmanned Aerial Vehicles and Deep Learning

Andreas Kamilaris[1] and Francesc X. Prenafeta-Boldú

Institute for Food and Agricultural Research and Technology (IRTA)

Barcelona, Spain

**Abstract:** Monitoring of disasters is crucial for mitigating their effects on the environment and human population, and can be facilitated by the use of unmanned aerial vehicles (UAV), equipped with camera sensors that produce aerial photos of the areas of interest. A modern technique for recognition of events based on aerial photos is deep learning. In this paper, we present the state of the art work related to the use of deep learning techniques for disaster identification. We demonstrate the potential of this technique in identifying disasters with high accuracy, by means of a relatively simple deep learning model. Based on a dataset of 544 images (containing disaster images such as fires, earthquakes, collapsed buildings, tsunami and flooding, as well as "non-disaster" scenes), our results show an accuracy of 91% achieved, indicating that deep learning, combined with UAV equipped with camera sensors, have the potential to predict disasters with high accuracy.

**Keywords**: Deep Learning, Disaster Monitoring, Unmanned Aerial Vehicles, Drones, Image Processing.

## 1. Introduction

Climate change and global warming lead to an increase of hydro-meteorological disasters around the world. According to the International Disaster Database (EM-DAT) (Centre for Research on the Epidemiology of Disasters, 2017), there is a large increase of natural and/or technological disasters in the last two decades, and this increase is expected to continue.

---

[1] Corresponding Author. Email: andreas.kamilaris@irta.cat

Although preparedness and prevention of disasters are processes far more important than dealing with them after happening (Paton, 2003), some are inevitable and cannot be easily predicted (Shaluf, Ahmadun, & Mat Said, 2003). Monitoring of possible disasters and fast identification of their occurrence are crucial for mitigating their effects on the physical environment or to humans. Fast alerting and immediate response are important aspects for dealing effectively with them by means of appropriate disaster management (Lettieri, Masella, & Radaelli, 2009).

Disaster monitoring can be facilitated by the use of unmanned aerial vehicles (UAV) such as drones, which are becoming more and more popular, having various benefits such as small size, low cost of operation, exposure to dangerous environments and high probability of mission success without the risk of loss of aircrew resource. These and other benefits are explained in related work, either independently (Petrides, Kolios, Kyrkou, Theocharides, & Panayiotou, 2017), (Zhang & Kovacs, 2012), or in relation to other remote sensing techniques such as airplanes and satellites (Matese, et al., 2015). UAV have demonstrated to compete with traditional acquisition platforms (i.e. satellite and aircraft) due to low operational costs, high operational flexibility and high spatial resolution of imagery.

UAV have been successfully used in scenarios of detecting earthquake-triggered roof-holes (Li, et al., 2015), oil spills and flooding (Luo, Nightingale, Asemota, & Grecos, 2015), victims (Andriluka, et al., 2010) etc. Drones are usually equipped with camera sensors, which could be optical cameras operating over the visible light spectrum, infrared to operate in low light conditions (by detecting emitted heat) or hyperspectral ones, considering particular spectra for events of interest. Images collected by the sensors can then be processed locally by an on-board computer system or sent to control centers for remote processing. In both cases of local and remote processing, *automatic* identification of events of interest (i.e. potential disasters) are important for sensing the disaster while it is happening, taking immediate life-saving actions, either by governmental organizations or the citizens themselves (Kamilaris & Pitsillides, 2014).

A modern, promising technique for image recognition is deep learning or convolutional neural networks (CNN) (LeCun, Bengio, & Hinton, 2015). Deep learning allows computational models composed of multiple processing layers to learn representations of data with multiple abstraction levels. Krizhevsky et al. demonstrated the potential of CNN in the ImageNet large scale visual recognition challenge (ILSVRC) in 2012 (Krizhevsky, Sutskever, & Hinton, 2012). Deep learning seems to have better performance in image recognition tasks (Liu & Wu, 2016), in comparison to common techniques used for analyzing images such as scalable vector machines (SVM) and artificial neural networks (ANN) (Saxena & Armstrong, 2014).

Thus, drones equipped with camera sensors using deep learning models to identify various (possibly disastrous) events constitute a recent practice with high potential (Zeggada, Melgani, & Bazi, 2017), as the identification is fast (i.e. in seconds after the model has learned the problem), flexible (i.e. many events can be identified in once) and scalable (i.e. learning process is continuous and adaptable).

**2. Related Work**

Related work involves disaster detection by using deep learning techniques. Table 1 lists the relevant efforts, together with their level of accuracy in respect to events' identification. From Table 1, the research efforts in #1 and #2 use aerial photos from UAV.

As Table 1 indicates, deep learning is a recent technique as all the efforts listed have been published in 2016-2017, while accuracy is relatively high (72-99%).

In this paper, our contribution ist two-fold:

- Presentation of the (still limited) work in this research field,
- Demonstration of the potential of deep learning and UAV for disaster monitoring based on a small dataset we have created (see Section 3), which contains various events of interest.

In the following section, we describe our methodology to develop this dataset, as well as the analysis performed and preliminary findings.

| No. | Disaster | Image source | Accuracy |
|-----|----------|--------------|----------|
| 1. | Fire (Kim, Lee, Park, Lee, & Lee, 2016) | Aerial photos | Human-like judgement |
| 2. | Avalanche (Bejiga, Zeggada, Nouffidj, & Melgani, 2017) | Aerial photos | 72-97% |
| 3. | Car accidents and fire (Kang & Choo, 2016) | CCTV cameras | 96-99% |
| 4. | Landslides (Liu & Wu, 2016) | Optical remote sensing | 96% |
| 5. | Landslides and flood (Amit, Shiraishi, Inoshita, & Aoki, 2016) | Optical remote sensing | 80-90% |

**Table 1:** Analysis of related work.

**3. Methodology**

As a deep learning model, we used the VGG architecture (Simonyan & Zisserman, 2014), pre-trained with the ImageNet[2] image dataset (Krizhevsky, Sutskever, & Hinton, 2012). VGG constitutes one of many successful" architectures, which researchers may use to start building their models instead of starting from scratch. Other popular ones include AlexNet (Krizhevsky, Sutskever, & Hinton, 2012), GoogleNet (Szegedy, et al., 2015) and Inception-ResNet (Szegedy, Ioffe, Vanhoucke, & Alemi, 2017). Each architecture has different advantages and scenarios where it is more appropriate to be used (Canziani, Paszke, & Culurciello, 2016). We selected VGG as it is easy to use, being automatically integrated to the TensorFlow open-source software library for machine intelligence as a Python class (TensorFlow, 2017).

---

[2] ImageNet Image Database. http://www.image-net.org/

Our aim was to "wire" the VGG architecture (i.e. the pre-trained convolutional layers) with a dense layer, trained with a dataset of aerial photos containing various natural disasters. As such dataset is not publicly available (to our knowledge), we decided to develop our own, by searching Google Images for relevant images, using the following keywords:

[Disaster | Landscape] + "aerial view" + "drone"

where [*Disaster*] could be *earthquake*, *hurricanes*, *flood* and *fire*. We selected only images originating from aerial views, most likely captured by UAV[3]. We decided to map together images from earthquakes and hurricanes showing collapsed buildings, as well as images from hurricanes, earthquakes or tsunami showing flood. For fire, we selected images showing active flames or smoke. Finally, we selected normal images of various [Landscapes] showing aerial views of *cities*, *villages*, *forests* and *rivers*, to examine if and how the deep learning model can actually predict disastrous events. The different groups of images are listed in Table 2.

| No. | Image Group | No. of Images | Relevant Possible Disaster |
|---|---|---|---|
| 1. | Buildings collapsed | 101 | Earthquakes and hurricanes |
| 2. | Flames or smoke | 111 | Fire |

---

[3] We read through some webpages (i.e. 10%) of the automatically identified images, to find out more if they were really taken from drones. The large majority was indeed captured by UAV.

| | | | |
|---|---|---|---|
| 3. | Flood | 125 | Earthquakes, hurricanes and tsunami |
| 4. | Forests and rivers | 104 | No Disaster |
| 5. | Cities, villages and urban landscapes | 103 | No Disaster |

**Table 2:** Description of our disaster dataset.

Based on this dataset, we trained our deep learning architecture to classify automatically aerial photos according to the image group (i.e. disaster category or landscape) where they belonged. To increase our dataset, we employed data augmentation techniques (Krizhevsky, Sutskever, & Hinton, 2012) to enlarge artificially the number of training images using label-preserving transformations, such as translations, transposing and reflections, and altering the intensities of the RGB channels. In this way, at every run of our training procedure (i.e. epoch), each image from the training dataset was randomly transformed before used as input to the model.

**4. Analysis**

In this section, we describe our findings by experimenting with our VGG architecture. We used 82% (444 images) of our dataset as training data and 18% (100 images) as testing data, dividing our dataset randomly in training and testing. We note that the combination 80-20 (i.e. for splitting training and testing data) showed the best results in our experiments (we also tried 70-30, 75-25, 85-15 and 90-10 with 3-6% less accuracy, see Figure 1). Besides, it is the combination highly preferred in the literature too (Deng & Yu, 2014), (Wan, et al., 2014).

We also note that the 91% precision was obtained after a total of 30 epochs, as shown in Figure 2. After this number of epochs, the precision was slightly decreased up to 88% at the 35th epoch. Hence, we stopped the learning procedure at the 30th epoch of the training. By epochs here, we refer to each complete run of the algorithm during training, using the whole of the training dataset each time.

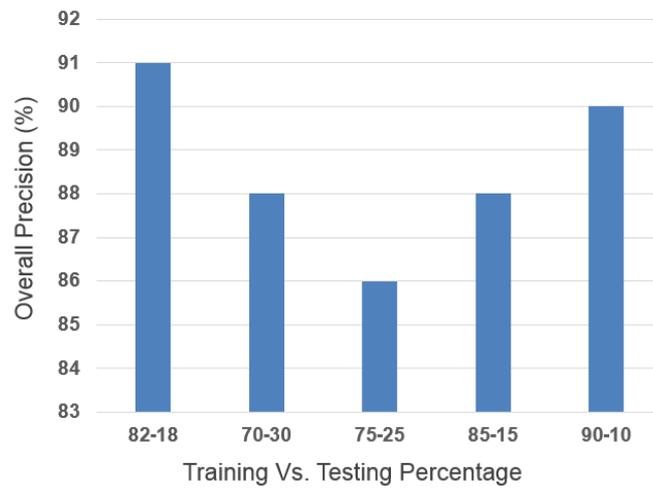

**Figure 1:** Precision vs. distribution of dataset in training/testing.

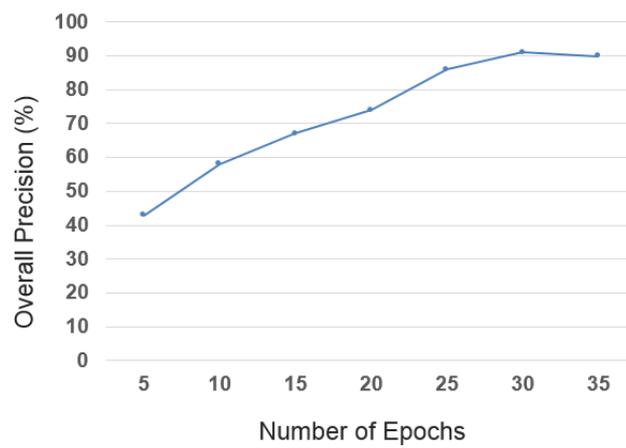

**Figure 2:** Precision vs. number of training epochs.

The training procedure required 20 minutes on a Linux machine (Intel Core i7, 8 MB RAM), while the testing procedure needed less than 5 minutes for all the 100 images used for testing. Our overall accuracy was 91% and the confusion matrix of the predictions vs. the actual classifications is depicted in Figure 3.

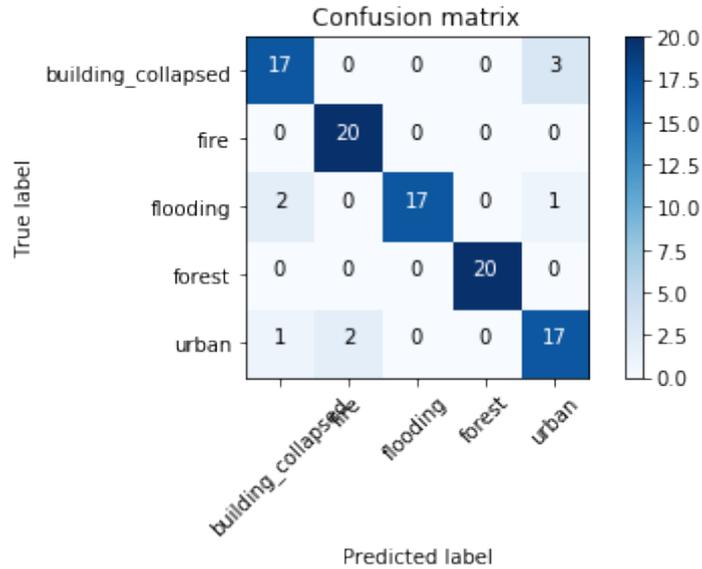

**Figure 3**: Confusion matrix of the predicted labels vs. the true ones.

It seems that the deep learning architecture managed to learn the problem relatively correctly, although it confused flooding events with urban landscapes that involved lakes and rivers or collapsed buildings (3% of the 9% total error) as well as buildings collapsed with urban landscapes (4% of the 9% total error), and vice-versa. Examples of this "confusion" are listed in Figure 4. An intuitive approach to reduce the error could be to avoid during training images of cities with rivers or lakes, or surfaces that reflect the sun, especially during sunset.

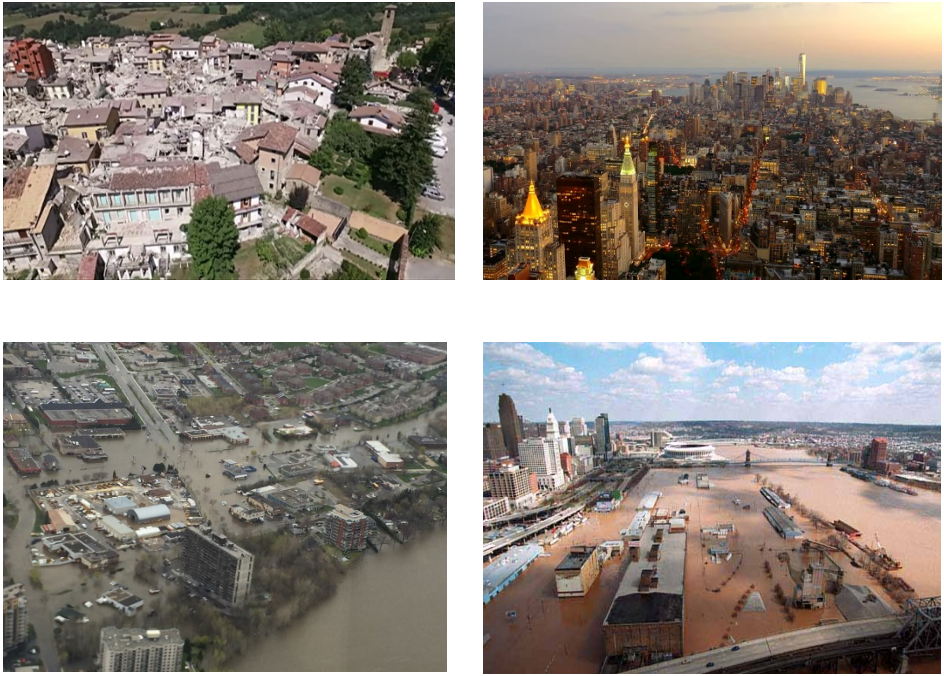

**Figure 4:** Examples of wrong predictions. Buildings collapsed instead of urban landscape (top-left). Fire instead of urban landscape (top-right). Buildings collapsed instead of flooding (bottom-left). Urban landscape instead of flooding (bottom-right).

**5. Discussion**

Analysis results in Section 4 have indicated the high accuracy obtained in disaster recognition by means of aerial photos, which could have been collected by means of cameras embedded in drones or other UAV.

Except from the relatively high accuracy, there are some other important advantages of using deep learning in image processing. Previously, traditional approaches for image classification tasks had been based on hand-engineered features, whose performance affected heavily the overall results. Feature engineering (FE) is a complex, time-consuming process which needs to be altered whenever the problem or the dataset changes. Thus, FE constitutes an expensive effort that depends on experts' knowledge and does not generalize well (Amara, Bouaziz, & Algergawy, 2017). On the other hand, deep learning does not require FE, locating the important features itself through training. It generalizes well (Pan & Yang, 2010) and it is quite robust even under challenging conditions such as illumination, complex background, different resolution, size and orientation of the images (Amara, Bouaziz, & Algergawy, 2017). Even though it takes sometimes considerable time to learn the problem, after it does, its testing time efficiency is much faster than other methods like SVM or KNN, even when considering hyperspectral cameras/images (Chen, Lin, Zhao, Wang, & Gu, 2014).

Its main disadvantages are that it takes (sometimes much) longer time to train, and it requires the preparation and pre-labeling of a large dataset, which would serve as the input during the training procedure. In spite of data augmentation techniques which augment some dataset with label-preserving transformations, in reality at least some hundreds of images are required, depending on the complexity of the problem under study (i.e. number of classes, precision required etc.). A problem with some datasets is the low variation among different classes (e.g. urban rivers/lakes vs. flooding) or the existence of noise in the form of low resolution, inaccuracy of sensory equipment, high sun/lights reflection in the image and others. Finally, although deep learning does not require FE, since data annotation is a necessary operation, some tasks are more complex and there is a need for experts (who might be difficult to involve) in order to annotate input images. In the case of disasters this is easier to do even by non-experts, but in general this is an important consideration.

## 6. Conclusion

In this paper, we have presented the state of the art work related to the use of deep learning techniques for disaster monitoring and identification, based on aerial photos captured by UAV. We have also created a small dataset of 544 images, creating a deep learning model based on VGG, in order to show the potential of this technique in identifying disasters automatically and with high accuracy. Based on our small dataset, by employing data augmentation techniques, accuracy has reached 91%. However, the authors expect that a larger dataset would effectively reduce the error (possibly reaching more than 95% accuracy) and that these results indicate that the CNN architecture used (and deep learning in general) has the potential to predict disasters with high accuracy in the near future.

Challenges in the use of UAV for disaster monitoring still exist, such as their limited computational capability, low energy resources and regulation issues for flight allowance, which hinder real-time data processing, area coverage and flexible use. Nevertheless, the opportunities combining UAV with deep learning techniques are large, as deep learning can provide high-accuracy event identification in real-time without requiring much processing capacity.

Finally, UAV with deep learning could enable better disaster modelling, especially when combined with geo-tagging of the events identified and geospatial applications. This would facilitate the integration of relevant actors (i.e. action forces/authorities, citizens/volunteers, other stakeholders) in disaster management activities with regard to communication, coordination and collaboration.

**Acknowledgments**

This research has been funded by the P-SPHERE project from the European Union's Horizon 2020 Research and Innovation Programme, under the Marie Skodowska-Curie grant agreement No 665919. The support of the CERCA Programme / Generalitat de Catalunya is also acknowledged.